\documentclass[runningheads]{llncs}
\usepackage[T1]{fontenc}
\usepackage{graphicx,verbatim}
\usepackage{amsmath}
\usepackage{amsfonts}
\usepackage{booktabs}
\usepackage{multirow}
\usepackage{nicefrac}
\usepackage{microtype}
\usepackage{xcolor}
\usepackage{url}
\usepackage{placeins}
\usepackage{multirow}
\usepackage{hyperref}
\hypersetup{hidelinks}
\usepackage{color}

\begin{document}
\title{{GLR-MM: Graph-Based Global--Local Reconstruction for Robust Multimodal Chest X-ray and EHR Representation Learning under Missing Modalities}}
\titlerunning{GLR-MM: Global-Local Reconstruction under Missing Modalities}
\author{Surbhi Sharma\inst{1}\textsuperscript{\textdagger} \and
Devesh Maheshwari\inst{2}\textsuperscript{\textdagger} \and
Nikhil Manali\inst{3}\textsuperscript{\textdaggerdbl}}
\authorrunning{S. Sharma et al.}
\institute{Houston Methodist, Houston, TX, USA\\
\email{ssharma3@houstonmethodist.org}\\
\email{shars06@pfw.edu}
\and
University of Wisconsin--Madison, Madison, WI, USA\\
\email{dmaheshwar22@wisc.edu}
\and
Walmart, USA\\
\email{nmanali@buffalo.edu}}

\maketitle              
\begingroup
\renewcommand{\thefootnote}{\textdagger}
\footnotetext{These authors contributed equally and share first authorship.}
\endgroup
\begingroup
\renewcommand{\thefootnote}{\textdaggerdbl}
\footnotetext{The last author provided project support.}
\endgroup
\begin{abstract}
Clinical multimodal models must often predict before all chest X-ray (CXR) and electronic health record (EHR) inputs are available. Existing approaches align observed representations, model missingness, or reconstruct across modalities, but do not jointly exploit within-patient and clinically similar inter-patient evidence. We propose \textbf{GLR-MM}, a Graph-Based Global--Local Reconstruction framework for early ICU mortality prediction. It maps five CXR--EHR modalities to a shared space, reconstructs missing embeddings through complementary local cross-modal and global graph-attention branches, adaptively fuses their estimates, and optimizes class-balanced prediction, reconstruction, and contrastive objectives.

On 9,620 MIMIC-derived ICU stays, we evaluate 10\%, 30\%, and 50\% random modality missingness with shared deterministic masks. MUSE performs better under mild and moderate missingness, whereas GLR-MM achieves higher AUROC and AUPRC at 50\% by 0.0088 and 0.0249, respectively. These results indicate that graph-guided reconstruction is most useful when inputs are severely incomplete.

\end{abstract}
\keywords{Multimodal Learning \and Missing Modalities \and Graph Neural Networks \and Clinical Prediction \and Representation Learning.}
\section{Introduction}
\label{sec:intro}

Clinical decision-making combines chest X-rays (CXR) with heterogeneous electronic health record (EHR) data, including laboratory measurements, physiological variables, diagnoses, and notes. These sources provide complementary views of patient state~\cite{huang2020fusion,shickel2018deep,baltrusaitis2018multimodal,ramachandram2017deep}, but arise from different workflows. Imaging, tests, or documentation may therefore be absent when an early risk estimate is required, exposing complete-input models to a substantially sparser deployment distribution.

Existing approaches use imputation, distillation, robust representation learning, missingness-aware fusion, graphs, or modality prompting~\cite{wang2020missing,ma2022smil,lee2023multimodal,wu2024muse,liang2025crlmmnar}. They generally align or fuse observed inputs, or model why an input is missing, rather than explicitly recovering each absent latent view from both the same patient and clinically similar patients.

We propose \textbf{GLR-MM}, a Graph-Based Global--Local Reconstruction framework. A local branch predicts a missing embedding from the patient's remaining views; a global branch retrieves the target modality from eligible neighbors in an observed-only patient graph. A learned gate combines both estimates, after which the completed modalities are fused and graph-refined for mortality prediction. Rebuilding neighborhoods only from observed inputs prevents reconstructions from influencing their own evidence.

We evaluate early mortality using five modalities from the first 24 hours of linked MIMIC-IV and MIMIC-CXR stays~\cite{johnson2023mimic,johnson2019mimiccxr}. Models share deterministic masks and a Monte Carlo evaluation bank at 10\%, 30\%, and 50\% missingness. Our contributions are: (1) reconstruction from intra- and inter-patient evidence; (2) adaptive fusion with observed-only graph construction; and (3) a controlled comparison showing MUSE stronger at mild and moderate missingness but GLR-MM strongest when half the modalities are unavailable.

\section{Related Work}
Clinical models fuse structured EHR variables, time series, imaging, and text because no single source captures the complete patient state~\cite{deasy2020dynamic,AdaCoMed}. Conventional fusion assumes a consistent input set, while zero replacement supplies no estimate of the information removed. Knowledge distillation transfers information from complete-input teachers~\cite{wang2020missing}; robust transformers tolerate arbitrary subsets~\cite{ma2022smil}; and prompting introduces learned tokens for absent inputs~\cite{lee2023multimodal}.

Graph and missingness-aware methods exploit additional structure. MUSE~\cite{wu2024muse} represents observations on a bipartite patient--modality graph and combines prediction with consistency objectives. CRL-MMNAR~\cite{liang2025crlmmnar} conditions prediction on observation patterns under non-random missingness. GLR-MM instead treats each absent modality as a recoverable latent view and supervises patient-specific local and cohort-derived global estimates against complete-input embeddings. Controlled MCAR corruption isolates this architectural distinction, although it does not reproduce every clinical workflow.

\section{Method}

\begin{figure}[t]
    \centering
    \includegraphics[width=\textwidth]{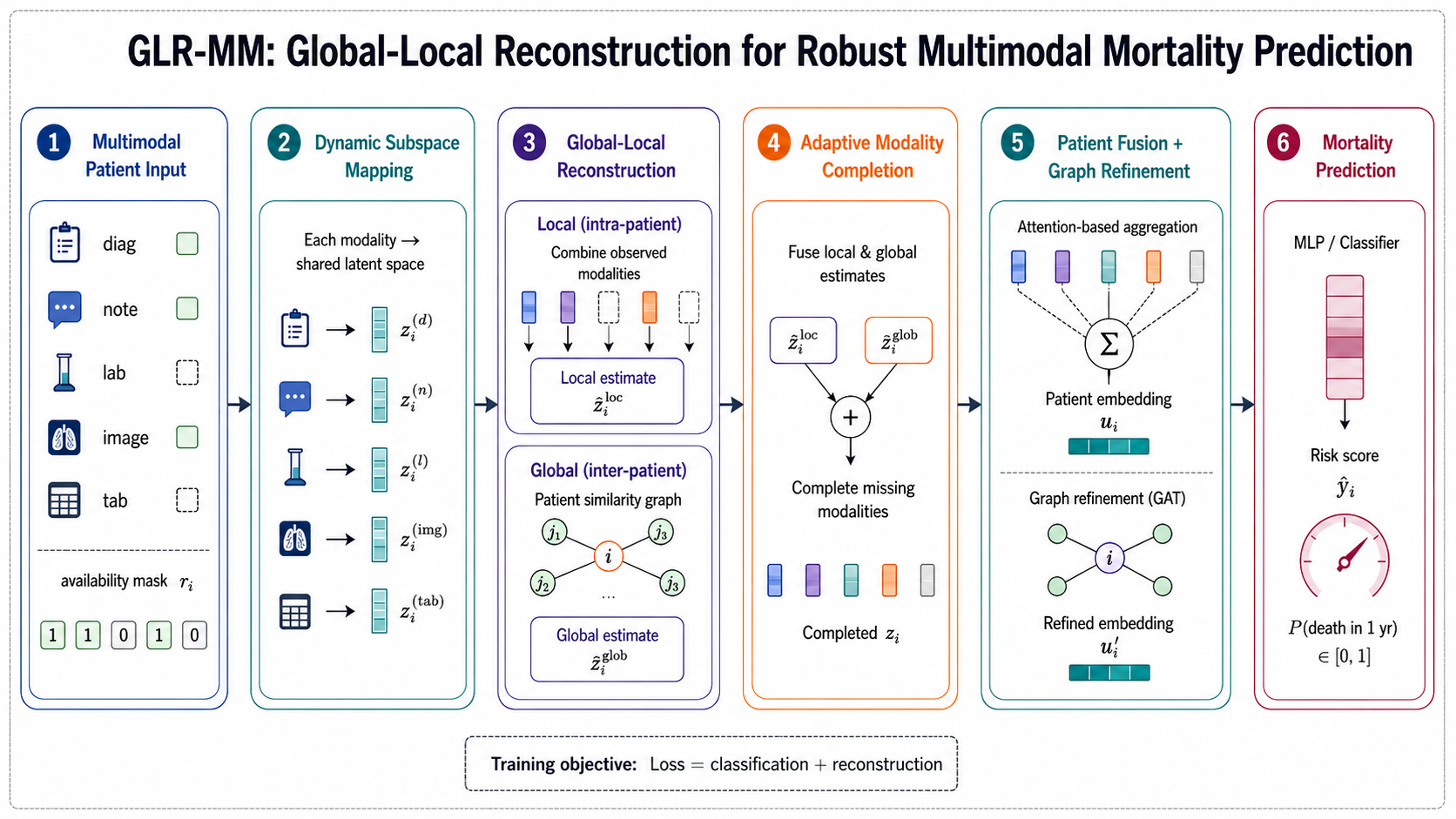}
    \caption{GLR-MM reconstructs five modality embeddings locally and globally, fuses them, and refines the patient representation on a graph for mortality prediction.}
    \label{fig:glrmm_overview}
\end{figure}

Patient $i$ has five modalities $x_i=\{x_i^{(m)}\}_{m\in\mathcal M}$, $\mathcal M=\{\mathrm{diag},\allowbreak\mathrm{note},\allowbreak\mathrm{lab},\allowbreak\mathrm{image},\allowbreak\mathrm{tab}\}$, mortality label $y_i\in\{0,1\}$, and availability mask $r_i\in\{0,1\}^{|\mathcal M|}$. GLR-MM reconstructs missing embeddings from within-patient and cohort evidence before graph-based prediction (Fig.~\ref{fig:glrmm_overview}). This separation is deliberate: reconstruction first restores a common multimodal representation, and the downstream predictor can then use one architecture for every availability pattern.

\subsection{Robust Multimodal Representation Learning}

\subsubsection{Dynamic modality encoding.}

Each modality is mapped to a common $d=128$ space. Structured variables first pass through a two-layer encoder; each modality-specific mapper combines a linear projection with an input-dependent mixture of four learned basis vectors:
\[
 \alpha_i^{(m)}=\operatorname{softmax}(g_m(x_i^{(m)})),\qquad
 z_i^{(m)}=\operatorname{LN}\!\left(W_mx_i^{(m)}+
 \alpha_i^{(m)}B_m\right),
\]
where $B_m\in\mathbb R^{4\times d}$ is the modality-specific basis bank. Separate mappers retain modality identity, while the common output dimension enables cross-modal reconstruction.

\subsubsection{Observed-only patient graph.}

For a given availability mask, we summarize only the observed modality embeddings,
\[
 \bar z_i=\frac{\sum_m r_i^{(m)}z_i^{(m)}}
 {\sum_m r_i^{(m)}},
\]
and construct a symmetrized Euclidean $k$-nearest-neighbor graph $G_r=(V,E_r)$ with $k=10$. It is rebuilt for every training and evaluation mask; reconstructed or propagated features never determine neighbors, so a missing target cannot affect its own neighborhood.

\subsubsection{Local reconstruction.}

The local branch predicts target $m$ from the patient's other observed modalities. With $\mathcal O_i=\{k:r_i^{(k)}=1\}$,

\[
 \beta_{ik}^{(m)}=\operatorname{softmax}_{k\in\mathcal O_i\setminus\{m\}}
 \!\left(s_m(z_i^{(k)})\right),\qquad
 \hat z_{i,\mathrm{loc}}^{(m)}=D_m^{\mathrm{loc}}
 \!\left(\sum_{k\in\mathcal O_i\setminus\{m\}}
 \beta_{ik}^{(m)}z_i^{(k)}\right).
\]
Excluding the target prevents an identity shortcut; a learned prior handles an empty context. Target-specific attention lets different observed views dominate different reconstructions.

\subsubsection{Global reconstruction.}

The global branch retrieves target-modality information only from neighbors that observe it. Let
$\mathcal N_i^{(m)}=\{j:(j,i)\in E_r,\ r_j^{(m)}=1\}$.
The patient summary provides a query, and eligible neighbors provide keys and values:

\[
 q_i^{(m)}=Q_m\bar z_i,\quad k_j^{(m)}=K_mz_j^{(m)},\quad
 v_j^{(m)}=V_mz_j^{(m)},
\]
\[
 \gamma_{ij}^{(m)}=\operatorname{softmax}_{j\in\mathcal N_i^{(m)}}
 \!\left(\frac{q_i^{(m)\top}k_j^{(m)}}{\sqrt d}\right),\qquad
 g_i^{(m)}=\sum_{j\in\mathcal N_i^{(m)}}\gamma_{ij}^{(m)}v_j^{(m)}.
\]
A context gate balances neighbor evidence and the patient query:
\[
 \eta_i^{(m)}=\sigma\!\left(a_m[q_i^{(m)};g_i^{(m)}]\right),\quad
 \hat z_{i,\mathrm{glob}}^{(m)}=D_m^{\mathrm{glob}}\!\left(
 \operatorname{LN}\!\left(\eta_i^{(m)}g_i^{(m)}+
 (1-\eta_i^{(m)})q_i^{(m)}\right)\right).
\]
If no neighbor is eligible, the branch uses a learned modality prior. Eligibility prevents recursive use of reconstructions, while the context gate can suppress a weak neighborhood.

\subsubsection{Adaptive fusion.}

The estimates are combined by a modality-specific gate,

\[
 \lambda_i^{(m)}=\sigma\!\left(h_m[\hat z_{i,\mathrm{loc}}^{(m)};
 \hat z_{i,\mathrm{glob}}^{(m)}]\right),\qquad
 \hat z_i^{(m)}=\lambda_i^{(m)}\hat z_{i,\mathrm{loc}}^{(m)}+
 (1-\lambda_i^{(m)})\hat z_{i,\mathrm{glob}}^{(m)}.
\]

Observed embeddings are retained, while missing ones are replaced by reconstructed embeddings,

\[
\tilde z_i^{(m)}=
\begin{cases}
z_i^{(m)}, & r_i^{(m)}=1,\\
\hat z_i^{(m)}, & r_i^{(m)}=0,
\end{cases}
\qquad
u_i=\sum_m a_i^{(m)}\tilde z_i^{(m)}.
\]
Here $a_i^{(m)}$ is attention over completed embeddings: $\lambda_i^{(m)}$ selects a reconstruction source, whereas $a_i^{(m)}$ controls its contribution to the patient representation.

\subsection{Graph-Based Outcome Prediction}

The observed-only graph is reused by a two-layer graph attention network with four heads then one head, residual connections, and layer normalization:
\[
 u_i'=\operatorname{GAT}_2\!\left(
 \operatorname{GAT}_1(u_i,G_r),G_r\right),\qquad
 \ell_i=c_\psi(u_i'),\qquad \hat y_i=\sigma(\ell_i).
\]

\subsection{Training Objective}

Each epoch uses a deterministic full-cohort MCAR mask keyed by seed, rate, and epoch while retaining one modality. With complete-input target $z_i^{\star(m)}$, branch $b\in\{\mathrm{loc},\mathrm{glob},\mathrm{fuse}\}$ incurs MSE only on removed modalities:

\[
 \mathcal L_b=\frac{\sum_{i,m}(1-r_i^{(m)})
 \lVert\hat z_{i,b}^{(m)}-z_i^{\star(m)}\rVert_2^2/d}
 {\sum_{i,m}(1-r_i^{(m)})}.
\]
\[
 \mathcal L_{\mathrm{rec}}=0.25\mathcal L_{\mathrm{loc}}+
 0.25\mathcal L_{\mathrm{glob}}+0.50\mathcal L_{\mathrm{fuse}}.
\]

An InfoNCE term $\mathcal L_{\mathrm{con}}$ (temperature 0.2) aligns jointly observed modalities from the same patient against other patients. The objective is

\[
 \mathcal L=\mathcal L_{\mathrm{cls}}+0.2\mathcal L_{\mathrm{rec}}+
 0.05\mathcal L_{\mathrm{con}},
\]
where $\mathcal L_{\mathrm{cls}}$ is binary cross-entropy weighted by $N_-/N_+$. Class weighting accounts for the 10.4\% mortality rate, reconstruction recovers missing modality representations, and self-supervised contrastive learning aligns co-observed modalities from the same patient while separating representations from different patients. Section~\ref{sec:experiments} gives training and evaluation details.

\section{Dataset}

We link MIMIC-IV EHR~\cite{johnson2023mimic}, MIMIC-CXR radiographs~\cite{johnson2019mimiccxr}, and MIMIC-IV-Note text~\cite{johnson2024mimicivnote} by patient and admission identifiers, retaining the first 24 hours for early in-hospital mortality prediction. Inputs are diagnosis, clinical-note, laboratory-text, CXR, and tabular/physiological representations. Text records are aggregated into fixed-length embeddings, and each radiograph is represented by a standardized image embedding. The tabular stream includes weight, monitoring availability, and \texttt{chartevents} summaries over 0--6 and 6--24 hours. The cohort contains \textbf{9,620 ICU stays}, including 1,002 deaths (10.4\%); preprocessing is fitted only on the training partition.

\section{Experiments}
\label{sec:experiments}

\subsection{Experimental Setup}
\label{sec:exp_setup}

We test robustness, comparison with missing-modality baselines, and the contributions of both reconstruction branches.

\subsubsection{Dataset and Missingness Protocol.}
We use a fixed stratified 72\%/8\%/20\% train/validation/test split and training-only standardization. MCAR corruption independently removes each modality at 10\%, 30\%, or 50\%, with one randomly selected modality restored whenever all five would otherwise be absent. This constraint makes every prediction well defined while retaining unstructured availability patterns.

All methods share deterministic training masks keyed by seed, rate, and epoch; MUSE slices the same cohort mask by mini-batch. Validation and test use ten shared masks per nonzero rate and one at 0\%. We average probabilities across each bank before computing AUROC and AUPRC, reducing dependence on a favorable corruption. AUPRC complements AUROC because mortality is uncommon.

\subsubsection{Training Protocol.}
Models train for at most 60 epochs with Adam ($5\times10^{-4}$ learning rate; $10^{-4}$ weight decay), unit-norm gradient clipping, plateau scheduling (factor 0.5; patience 3), and early stopping (patience 10) on validation AUPRC. GLR-MM uses $\lambda_{\mathrm{rec}}=0.2$, $\lambda_{\mathrm{con}}=0.05$, and $k=10$. Models trained at each nonzero rate are evaluated at 0\%, 10\%, 30\%, and 50\% over seeds 11, 22, 33, 44, and 55; we report mean and standard deviation.

\subsubsection{Implementation Details.}
GLR-MM uses hidden dimension 128, dropout 0.1, four bases per modality, two GAT layers (four heads then one), and two-layer tabular and classifier MLPs. MUSE uses batches of 256; cohort-dependent GLR-MM and CRL-MMNAR use full-cohort updates. Experiments run on one NVIDIA GPU. Code is available at \href{https://github.com/Devesh-Maheshwari/GLR-MM}{\textcolor{blue}{\underline{\nolinkurl{https://github.com/Devesh-Maheshwari/GLR-MM}}}}.

\subsection{Baselines and Component Ablations}
\label{sec:baselines}


\textbf{MUSE}~\cite{wu2024muse} encodes observed features on a bipartite patient--modality graph with two edge-aware GraphSAGE layers. We retain its prediction and mutual-consistency losses, 15\% edge dropout, temperature 0.05, and projection head. It uses the same mask bank and probability-level Monte Carlo ensemble as GLR-MM.

\textbf{CRL-MMNAR}~\cite{liang2025crlmmnar} conditions fusion on observation patterns and combines mortality, mask-prediction, leave-one-out reconstruction, and contrastive losses (weights 1.0, 0.5, 1.0, 0.3). Its two-fold pattern-wise residual rectifier is fitted only on validation predictions. Ablations retain local or global reconstruction; all variants otherwise share encoders, downstream GAT, masks, and optimization.
\subsection{Results}
\label{sec:results}

\subsubsection{Main robustness results.}
Table~\ref{tab:glrmm_robustness_compact} shows that GLR-MM's MC-ensembled AUROC spans 0.8537--0.8760 and AUPRC 0.4798--0.5661. AUPRC declines more than AUROC as evaluation corruption grows. At 50\% evaluation missingness, training at 50\% rather than 10\% raises AUROC from 0.8537 to 0.8730 and AUPRC from 0.4798 to 0.5304. The 50\%-trained model also has the strongest complete-input scores, consistent with a regularizing effect from modality dropout.

The cross-rate grid also measures sensitivity to a training--deployment mismatch. From complete evaluation to 50\% missingness, the 10\%-trained model loses 0.0070 AUROC and 0.0558 AUPRC, whereas the 50\%-trained model loses 0.0030 and 0.0357. Exposure to more incomplete patterns during optimization therefore reduces degradation across the evaluation range. This does not prove that 50\% corruption is optimal for every cohort, but it shows that the assumed training availability should reflect the intended deployment setting.

\begin{table}[t]
\centering
\small
\caption{\textbf{GLR-MM robustness.} Mean $\pm$ standard deviation across seeds.}
\label{tab:glrmm_robustness_compact}
\begin{tabular}{llcccc}
\hline
\textbf{Train} & \textbf{Metric} & \textbf{Eval 0\%} & \textbf{Eval 10\%} & \textbf{Eval 30\%} & \textbf{Eval 50\%} \\
\hline
\multirow{2}{*}{10\%} & AUROC & $0.861\pm.006$ & $0.859\pm.007$ & $0.858\pm.004$ & $0.854\pm.004$ \\
& AUPRC & $0.536\pm.008$ & $0.521\pm.005$ & $0.495\pm.006$ & $0.480\pm.009$ \\
\multirow{2}{*}{30\%} & AUROC & $0.869\pm.008$ & $0.866\pm.008$ & $0.865\pm.008$ & $0.862\pm.009$ \\
& AUPRC & $0.546\pm.035$ & $0.530\pm.035$ & $0.515\pm.034$ & $0.502\pm.044$ \\
\multirow{2}{*}{50\%} & AUROC & $0.876\pm.004$ & $0.876\pm.002$ & $0.873\pm.002$ & $0.873\pm.002$ \\
& AUPRC & $0.566\pm.010$ & $0.553\pm.011$ & $0.535\pm.010$ & $0.530\pm.009$ \\
\hline
\end{tabular}
\end{table}

\subsubsection{Comparison to baselines.}



At matched rates (Table~\ref{tab:evaluation_missing_all}), MUSE leads GLR-MM at 10\% by 0.0146 AUROC/0.0284 AUPRC and at 30\% by 0.0020/0.0099. Thus, when most modalities remain present, direct graph representation learning is sufficient and explicit reconstruction does not improve the aggregate scores. The gap narrows at 30\%, indicating that the two strategies behave similarly under moderate loss.

At 50\%, the ordering reverses: GLR-MM leads MUSE by 0.0088 AUROC and 0.0249 AUPRC. This crossover supports the intended use of global--local reconstruction, because the opportunity to recover absent information grows as the observed patient representation becomes sparse. GLR-MM also exceeds adapted CRL-MMNAR at all three matched rates, although the latter has large seed-to-seed variation. Because architectures differ and no paired hypothesis test was performed, these are descriptive rather than statistically established differences.

Accordingly, the comparison does not support one architecture for every operating condition. MUSE is preferable in the observed mild and moderate regimes, while GLR-MM is preferable in the severe regime. A real deployment should first estimate its modality-availability distribution and then validate the selected model under those patterns, rather than choosing from complete-input performance alone.
\begin{table}[t]
\centering
\small
\caption{\textbf{Performance at matched training and evaluation missingness rates.} Results are reported as mean $\pm$ standard deviation across five seeds.}
\label{tab:evaluation_missing_all}
\begin{tabular}{llccc}
\hline
\textbf{Model} & \textbf{Metric} & \textbf{10\%} & \textbf{30\%} & \textbf{50\%} \\
\hline

\multirow{2}{*}{GLR-MM}
& AUROC & 0.859 $\pm$ 0.007 & 0.865 $\pm$ 0.008 & $\mathbf{0.873 \pm 0.002}$ \\
& AUPRC & 0.521 $\pm$ 0.005 & 0.515 $\pm$ 0.034 & $\mathbf{0.530 \pm 0.009}$ \\
\hline

\multirow{2}{*}{MUSE}
& AUROC & $\mathbf{0.873 \pm 0.004}$ & $\mathbf{0.867 \pm 0.009}$ & 0.864 $\pm$ 0.008 \\
& AUPRC & $\mathbf{0.549 \pm 0.003}$ & $\mathbf{0.525 \pm 0.026}$ & 0.505 $\pm$ 0.013 \\
\hline

\multirow{2}{*}{CRL-MMNAR}
& AUROC & 0.794 $\pm$ 0.126 & 0.796 $\pm$ 0.133 & 0.735 $\pm$ 0.159 \\
& AUPRC & 0.430 $\pm$ 0.161 & 0.417 $\pm$ 0.155 & 0.335 $\pm$ 0.179 \\
\hline

\end{tabular}
\end{table}

\subsubsection{Component ablation.}
The paired ablation (Table~\ref{tab:glrmm_ablation}) yields nearly identical AUROC across Full, Local, and Global variants. At 10\%, the full model is marginally strongest on both metrics, while the Global branch has the highest 30\% AUROC by 0.0008. These differences are small relative to the reported standard deviations.

The clearest effect appears in AUPRC. Fusion gives the strongest AUPRC at 30\% and 50\%; at 50\%, it improves by 0.0081 over Local and 0.0068 over Global. Neither branch therefore dominates across all conditions. Their adaptive combination mainly benefits precision--recall behavior in the regime where reconstruction is most frequently required.

\begin{table}[t]
\centering
\small
\caption{\textbf{Component ablation at matched missingness.} Mean $\pm$ standard deviation across five seeds.}
\label{tab:glrmm_ablation}
\begin{tabular}{llccc}
\hline
\textbf{Variant} & \textbf{Metric} & \textbf{10\%} & \textbf{30\%} & \textbf{50\%} \\
\hline
\multirow{2}{*}{Full}
& AUROC & $\mathbf{0.8588 \pm 0.0037}$ & 0.8658 $\pm$ 0.0074 & $\mathbf{0.8706 \pm 0.0018}$ \\
& AUPRC & $\mathbf{0.5186 \pm 0.0180}$ & $\mathbf{0.5270 \pm 0.0155}$ & $\mathbf{0.5278 \pm 0.0111}$ \\
\hline
\multirow{2}{*}{Local}
& AUROC & 0.8585 $\pm$ 0.0058 & 0.8658 $\pm$ 0.0060 & 0.8702 $\pm$ 0.0004 \\
& AUPRC & 0.5164 $\pm$ 0.0222 & 0.5250 $\pm$ 0.0089 & 0.5198 $\pm$ 0.0105 \\
\hline
\multirow{2}{*}{Global}
& AUROC & 0.8585 $\pm$ 0.0038 & $\mathbf{0.8666 \pm 0.0053}$ & 0.8705 $\pm$ 0.0035 \\
& AUPRC & 0.5178 $\pm$ 0.0177 & 0.5249 $\pm$ 0.0164 & 0.5211 $\pm$ 0.0172 \\
\hline
\end{tabular}
\end{table}
\FloatBarrier

\section{Conclusion}

We introduced GLR-MM, a multimodal framework that treats unavailable clinical modalities as recoverable latent views. Local reconstruction uses the patient's remaining inputs, global reconstruction retrieves target-modality evidence from observed-only patient neighborhoods, and adaptive fusion combines the estimates before outcome prediction.

Under a shared missingness and Monte Carlo evaluation protocol, MUSE leads at 10\% and 30\% matched missingness, whereas GLR-MM leads by 0.0088 AUROC and 0.0249 AUPRC at 50\%. The result identifies a practical tradeoff rather than universal dominance: explicit reconstruction becomes most useful as available evidence becomes sparse. Ablations show similar AUROC for either branch and the clearest fusion gain in AUPRC at 50\%, motivating further study of reconstruction for severely incomplete inputs.

\section{Limitations}

Evaluation uses one MIMIC-derived cohort, so transportability to other hospitals, acquisition protocols, and patient populations remains unknown. The simulated MCAR masks support a controlled architectural comparison but do not capture clinically informative missingness caused by severity, workflow, resources, or clinician decisions. Natural availability patterns may therefore change both absolute performance and the relative ordering of methods.

Patient neighborhoods may also become unreliable when few modalities remain, and an incorrect neighbor can introduce noise into a global reconstruction. The present results do not quantify graph uncertainty or test whether reconstructed embeddings are clinically faithful beyond their downstream utility. Future work will examine external cohorts, non-random missingness, calibrated uncertainty, and safeguards that abstain from reconstruction when local or cohort evidence is insufficient.


\begin{credits}
\subsubsection{\discintname}
The authors have no competing interests to declare that are relevant to the content of this article.
\end{credits}

%
%
%
\bibliographystyle{splncs04}
\bibliography{main}

\end{document}